\documentclass[sigconf,nonacm]{acmart}

\makeatletter
\fancypagestyle{firstpagestyle}{%
  \fancyhf{}%
  \fancyfoot[L]{\footnotesize
    Workshop on AI in Finance (co-located with ICAIF 2025), Singapore — November 2025.}%
  \fancyfoot[R]{\thepage}%
}
\makeatother
\AtBeginDocument{%
  }

\usepackage{tikz}
\usetikzlibrary{positioning, arrows.meta, fit}
\usepackage{cleveref}
\usepackage{subcaption}
\usepackage{booktabs}
\usepackage{multirow}
\usepackage{makecell}
\begin{document}

\title{Robust Yield Curve Estimation for Mortgage Bonds \\ Using Neural Networks}

 \author{Sina Molavipour, Alireza M. Javid, Cassie Ye, Björn Löfdahl, Mikhail Nechaev}
 \email{{sina.molavipour,alireza.javid,cassie.ye,bjorn.lofdahl,mikhail.nechaev}@seb.se}
\authornotemark[1]
 \affiliation{%
   \institution{SEB Group, Stockholm, Sweden}
   \country{}
 }


\begin{abstract}
Robust yield curve estimation is crucial in fixed-income markets for accurate instrument pricing, effective risk management, and informed trading strategies. Traditional approaches, including the bootstrapping method and parametric Nelson-Siegel models, often struggle with overfitting or instability issues, especially when underlying bonds are sparse, bond prices are volatile, or contain hard-to-remove noise. In this paper, we propose a neural network-based framework for robust yield curve estimation tailored to small mortgage bond markets. Our model estimates the yield curve independently for each day and introduces a new loss function to enforce smoothness and stability, addressing challenges associated with limited and noisy data. Empirical results on Swedish mortgage bonds demonstrate that our approach delivers more robust and stable yield curve estimates compared to existing methods such as Nelson-Siegel-Svensson (NSS) and Kernel-Ridge (KR). Furthermore, the framework allows for the integration of domain-specific constraints, such as alignment with risk-free benchmarks, enabling practitioners to balance the trade-off between smoothness and accuracy according to their needs.
\end{abstract}

\keywords{Yield curve estimation, mortgage bond, neural network}

\maketitle
\newcommand\blfootnote[1]{%
  \begingroup
  \renewcommand\thefootnote{}\footnote{#1}%
  \addtocounter{footnote}{-1}%
  \endgroup
}

\blfootnote{\footnotesize Workshop on AI Meets Quantitative Finance (held within ICAIF 2025, Singapore, November 2025)}

\section{Introduction}
The yield curve is a fundamental building block that underpins the pricing, valuation, and risk measurement of a broad spectrum of financial instruments, including bonds, FRNs, repos, and various structured products. Accurate estimation of the term structure of interest rates, commonly referred to as the yield curve, holds paramount importance for a wide range of stakeholders, including investors, policymakers, and financial institutions. In risk management, yield curves provide essential input for calculating various risk measurements, such as sensitivities and value-at-risk. Different central banks that utilize yield curve information employ Yield Curve Control (YCC) to sell or buy bonds, thereby maintaining the long-term interest rate at the target level to stimulate investments, support the economy, and control inflation. Short-term treasury yields are reflecting market expectations of central banks' policy changes, such as rate cuts or hikes. Traders rely on yield curve information to decide on trading strategies, such as riding the yield curve to profit from the upward slope in a stable interest rate environment. Yield curves also reflect the overall market condition and expectation. The 10Y$-$2Y Treasury yield spread is an indicator of overall market expectations. An inverted yield curve may indicate expectations of lower future interest rates or a potential slowdown in future growth. While yield curve segments range from treasury yields and corporate bond yields to mortgage or covered bond yields, the overall estimation techniques can be horizontally applied regardless of the segments. The yield curve represents spot rates (current market yield) for bonds of different maturities, which can be estimated for a given set of bonds of similar features within a segment. The prominent influence of the yield curve in finance and economics suggests that any inaccuracies in its estimation can propagate into significant mis-pricings, suboptimal risk and trading management strategies, and potentially flawed monetary policy decisions. This establishes a high standard for model accuracy and robustness, thereby motivating the continuous pursuit of advanced estimation methodologies. Analyzing and calibrating the estimated curves and rates are among the main daily routines in financial institutions. It is especially challenging for smaller market caps, with issuances often concentrated around short to mid-term periods and sparser over the long term, such as Swedish covered bonds.  

The estimation of the yield curve has evolved, shifting from flexible, non-parametric methods to more structured, parametric models. Early techniques often relied on spline-based methods—such as quadratic, cubic, exponential, and B-splines \cite{McCulloch-1971, McCulloch-1975}, which provided flexibility and could fit observed data well. However, these methods often led to unstable or irregular shapes, especially at the short and long ends of the curve \cite{Nelson-Siegel}. To address these issues, the Nelson-Siegel (NS) model was introduced in \cite{Nelson-Siegel}, offering a simple parametric form that aimed to capture key properties of a well-behaved yield curve, including smoothness, continuity, and the ability to represent both level and slope changes. Svensson later extended this model by adding more flexibility to the curve's shape \cite{svensson1994estimating}, which is referred to as the Nelson-Siegel-Svensson (NSS) model. Later on, the dynamic Nelson-Siegel (DNS) model extends this framework by modeling the evolution of the yield curve's underlying factors over time, enabling forecasting and capturing the temporal dynamics of interest rates based on historical data \cite{diebold2006DNS, diebold2008DNS}. In contrast to the NSS method, functional approximation can be achieved through a linear combination of kernel functions and weights, where the functions are determined by solving an error loss function based on bond prices or yield rates. In a recent paper, authors in~\cite{Filipovic2022stripping} introduce a kernel ridge (KR) model and show a closed-form solution by introducing a regularized loss incorporating smoothness of the curve, and argue that the estimates outperform existing parametric and non-parametric methods.

From a machine learning standpoint, estimating the yield curve from bond data can be treated as a functional approximation problem, where feedforward neural networks are known to be effective \cite{Hornik1989MultilayerFN}. While neural networks (NN) have been previously used for forecasting the yield curve over time ~\cite{kauffmann2022learning, Tarek2025}, for example by extending the dynamic Nelson-Siegel (DNS) framework, we use neural networks to model the yield curve independently for each day, without relying on temporal dependency across days, as a direct extension of NSS and KR models which has not been attempted in the literature. We demonstrate that core properties of the yield curve—such as smoothness and stability—can be enforced through the design of a novel loss function during training. Our main contributions are:
\begin{enumerate}
\item We demonstrate that our neural network-based model provides a more robust yield curve estimate compared to the existing methods, such as NSS and KR.
\item Our results demonstrate improved stability and reduced sensitivity to noise or fluctuations in bond prices, particularly in a small-data setting such as the Swedish mortgage bond market.
\item Our novel loss function enables the integration of domain-specific constraints (e.g., alignment with risk-free benchmarks), while balancing the trade-offs between accuracy and smoothness.
\end{enumerate}

In Section~\ref{sec:preliminaries}, we define the problem of yield curve estimation based on a set of underlying bonds and review several standard estimation techniques.
Section~\ref{sec:neural_net} presents our proposed neural network architecture and the corresponding loss functions used to regularize training.
In Section~\ref{sec:experiment}, we describe our experimental setup. We begin with hyperparameter tuning, followed by an evaluation of our model's performance in terms of robustness to outliers, day-to-day stability, and the trade-off between smoothness and flexibility in a leave-one-out setup.
Finally, Section~\ref{sec:conclusion} summarizes our findings and outlines potential directions for future research.

\section{Preliminaries and related works}
\label{sec:preliminaries}
In this section, we cover some of the known methods for estimating the yield curve. Let $y(t)$ be the spot yield rate at maturity time $t$, commonly in years. Let $f(t)$ denote the forward curve at maturity $t$. The yield rate can then be computed as:
\begin{equation}
\label{eq:forward_yield}
    y(t)=\frac{1}{t}\int_0^t f(\tau)d\tau.
\end{equation}
In order to estimate the present value of a bond, the face-value and future cashflow payments must be discounted to the present time. Assuming a given yield curve $y(t)$, the discount factor at time $t$ can be calculated as:
\begin{equation}
\label{eq:discount-yield}
    d(t)=e^{-ty(t)},
\end{equation}
where we use the notion of continuous compounding, although market practices may vary depending on the instrument. Consider a dataset of $M$ bonds sold in the market on a given day. The present value of the bond $j$ with $n_j$ periodic cashflows can be estimated as:
\begin{equation}
\label{eq:present_value}
    \hat{p}_j = \sum_{i=1}^{n_j-1}c^{(i)}_j d\left(t^{(i)}_j\right) + \left(c^{(n_j)}_j + F_j\right)d\left(T_j\right), \qquad  j\in{1,\dots,M}.
\end{equation}
where $t_j^{(i)}$s are cashflow dates (in years), $c_j^{(i)}$ is the cashflow amount at time $t_j^{(i)}$, and $F_j$ is the face-value of the bond maturing at $T_j=t_j^{(n_j)}$. The present value of the bond can then be compared with its currently observed market price $p_j$ to evaluate the estimation accuracy of the yield curve $y(t)$. In other words, we would want to have $\sum_{j=1}^M (p_j-\hat{p}_j)=0$ in an ideal situation. Estimating the yield curve, or equivalently, the discount curve, for this set of equations is non-trivial, and various approaches can be employed. According to (\ref{eq:present_value}), and the number of bonds observed in the market $M$, the problem is under-determined due to having discrete observations for a continuous function, resulting in a non-smooth curve that aims to satisfy this set of constraints. The resulting discrete points require interpolation to create a continuous curve. However, naive interpolation can produce forward curves with negative rates or excessive volatility since we have $f(t) = y(t) + t\frac{dy(t)}{dt}$.

One of the most fundamental techniques for constructing a zero-coupon yield curve is the bootstrapping method, which enables practitioners to derive appropriate discount rates from observable market bond quotes. The method incrementally builds the yield curve by solving for the implied spot rates sequentially, starting with the bond of the shortest maturity and then using that solution to solve for the bond with the second shortest maturity, and so forth. This recursive structure makes bootstrapping particularly robust when a complete set of liquid bond instruments exists across the desired maturity spectrum. However, real-world limitations such as non-uniform maturities, pricing errors, and liquidity constraints can make the process sensitive to data quality and interpolation methods.

The bootstrapping method is widely used in practice for constructing term structures of interest rates, particularly for risk-free rates such as those derived from government securities or Overnight Index Swaps (OIS). While intuitive and relatively easy to implement, bootstrapping does not enforce smoothness across the curve, which can lead to local irregularities unless post-processing or interpolation (e.g., spline fitting) is applied.

\subsection{Nelson-Siegel-Svensson}
Nelson-Siegel \cite{Nelson-Siegel} used a parsimonious parametric functional form to model the forward rate:
\begin{equation}
    f_{NS}(t)=\beta_0+\beta_1 e^{-\frac{t}{\lambda}} + \beta_2 \frac{t}{\lambda} e^{-\frac{t}{\lambda}},    
\end{equation}
and accordingly, the yield curve is obtained using (\ref{eq:forward_yield}):
\begin{equation}
    y_{\text{NS}}(t) = \beta_0 + \beta_1 \frac{1 - e^{-t/\lambda}}{t/\lambda} + \beta_2 \left( \frac{1 - e^{-t/\lambda}}{t/\lambda} - e^{-t/\lambda} \right).    
\end{equation}
The motivation for this parametric model was to capture the common shapes of the yield curve, including monotonic forms and extreme points in specific parts of the curve. Later, more terms were added to the model by Svensson \cite{svensson1994estimating} to capture more complex behavior in the rates:
\begin{equation}
    f_{\text{NSS}}(t)=\beta_0 + \beta_1 e^{-t/\lambda_1} + \beta_2 \left( \frac{t}{\lambda_1} \cdot e^{-t/\lambda_1} \right) + \beta_3 \left( \frac{t}{\lambda_2} e^{-t/\lambda_2} \right),    
\end{equation}
which results in:
\begin{align}
    y_{\text{NSS}}(t)=&\beta_0 + \beta_1 \frac{1 - e^{-t/\lambda_1}}{t/\lambda_1} + \beta_2 \left( \frac{1 - e^{-t/\lambda_1}}{t/\lambda_1} - e^{-t/\lambda_1} \right) \\
    &+ \beta_3 \left( \frac{1 - e^{-t/\lambda_2}}{t/\lambda_2} - e^{-t/\lambda_2} \right).    
\end{align}
Although this model has been extensively applied in finance and banking \cite{ECB2018,canada1999}, it exhibits several limitations. A well-known issue is the lack of robustness in its estimations. In practical applications, the bond dataset often undergoes cleaning, with bonds being added or removed depending on market conditions. As a result, the estimated yield curves derived from these parsimonious models can vary significantly, particularly in the short/long end of the curve.

\subsection{Kernel-ridge method}
In this approach, the discount function is modeled by kernel functions. In a recent work \cite{Filipovic2022stripping}, the authors show that there is a unique closed-form solution when using this model to optimize the price error while incorporating smoothness conditions in the objective function:
\begin{equation}
    \min_{d} \textstyle\sum_{j=1}^M \omega_j (p_j-\hat{p}_j)^2 + \lambda ||d||^2 ,
\end{equation}
for some smoothness parameter $\lambda>0$, where the norm in the second term is defined by the weighted average of the first and second derivative of $d(\cdot)$ to ensure the smoothness (see \cite{Filipovic2022stripping}).
Then, by writing the kernel representation for the discount function as below, the optimization problem can be solved:
\begin{equation}
    \hat{d}(t) = 1+\textstyle\sum_{l=1}^{L} \alpha_l k(t,t_l).
\end{equation}
$k(t,t_l)$ are kernel functions that form a RKHS (reproducing kernel Hilbert space). So, the corresponding kernel matrix $K$ is constructed by $K_{ml}=k(t_m,t_l)$. The closed-form solution determines both the weights $\alpha_l$ and the $K$ based on boundary conditions on the smoothness criteria. In this paper, we refer to this method as KR where we set the weights $\omega_j$ as inversely proportional to the squared duration $D_j$, that is $\omega_j = \frac{1}{M}\frac{1}{(D_{j} p_{j})^2}$, which approximates the mean squared yield fitting error as stated in \cite{Filipovic2022stripping}. We found that this choice of $\omega_j$ results in a smoother yield curve that is less sensitive to sporadic price changes in small-sized markets such as mortgage bonds.

\section{Neural network estimation}
\label{sec:neural_net}
Neural networks are well-studied methods in machine learning and are widely used to estimate complex models due to their approximation power. In this paper, we investigate how neural networks can be used to estimate the yield curve $y(t)$, and tailor the objective function to reflect more complex criteria on the obtained curve. We argue that the main advantage of using neural networks is robustness in estimation and their flexibility to handle extra criteria, such as maintaining the economical reasonableness of the curve in various market conditions.

Since the mortgage bond market is commonly less liquid and populated than other markets (particularly in Sweden), the available data for training is relatively limited. Consequently, we adopt simple feed-forward neural network architectures with shallow layers and a small number of neurons to ensure effective parameter training. The quest for the most suitable architecture to achieve the best estimation accuracy is out of the scope of this paper. For a given activation function $\phi(.)$, our model takes the maturity time $t$ as input and produces the estimated yield rate $\hat{y}(t)$ at the output layer:
\begin{equation}
\label{eq:NN_model}
    \hat{y}(t)=\textstyle\sum_{i=1}^H v_i \phi(w_i\cdot t + b_i) + c,
\end{equation}
where $H$ is the number of hidden neurons.

\subsection{Loss function}
There are various ways to explain a ``good'' yield curve. When constructing yield curves for mortgage-backed securities, analysts must strike a delicate balance between market accuracy and economic plausibility. This trade-off arises from two competing objectives:
\begin{enumerate}
    \item \textbf{Accuracy}: The curve must precisely replicate observed market prices of mortgage bonds to ensure valid risk management calculations and hedge effectiveness. This can also be seen by comparing the yield values with the yield-to-maturity (YTM) of the underlying bonds. 
    \item \textbf{Economic Reasonableness}: The curve must maintain a logical relationship in terms of smoothness and trend with risk-free benchmarks.
\end{enumerate}

The comparison between the present value of bonds and their market prices is a well-established technique for addressing the accuracy of the estimation. This can be reflected in:
\begin{equation}
\label{eq:loss-1}
    \mathcal{L}_{\text{error}} = \frac{1}{M}\textstyle\sum_{j=1}^M (p_j-\hat{p}_j)^2.
\end{equation}
Additional criteria can guide the estimation toward a more well-behaved yield curve. 

A key aspect of conventional methods is the use of relatively smooth functions to model the yield curve. This smoothness can be enforced in the objective function by incorporating derivative terms of the estimated yield curve.
Consider a set of $N$ ordered fixed maturity grid points $\vec{t}=[t_1,\dots,t_N]$. Then, by computing the slope of the estimated curve at these points, we add the following penalty term to (\ref{eq:loss-1}):
\begin{equation}
    \mathcal{L}_{\text{smooth}} = \max_{i=2,\dots,N} |\frac{\hat{y}(t_i) - \hat{y}(t_{i-1})}{t_i - t_{i-1}}|.
\end{equation}

\begin{figure*}[htbp]
    \centering
    \includegraphics[width=\linewidth]{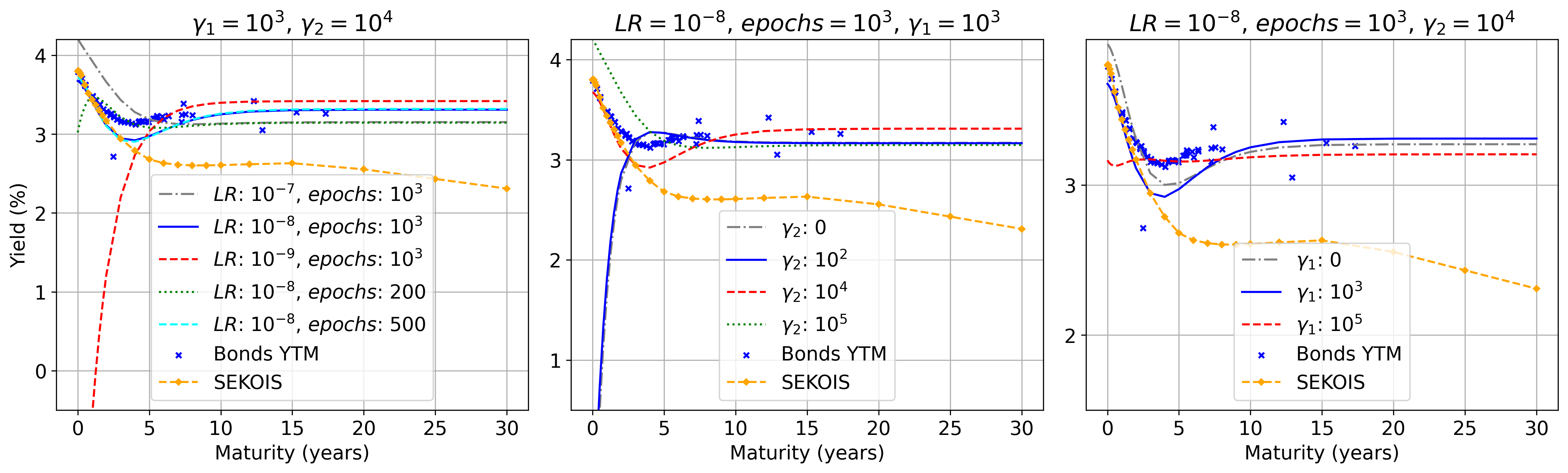}
    \caption{Hyperparameter tuning for learning rate (LR), number of epochs, $\gamma_1$, and $\gamma_2$ in a falling market (3/6/2024). Left: varying LR and epochs. Center: varying $\gamma_2$. Right: varying $\gamma_1$.}
    \label{fig:hyperparam-merged}
\end{figure*}

Another criterion to consider is the economical reasonableness of the estimated curve in various market conditions. For instance, to evaluate the risk premium of a mortgage bond, it is common practice to compare the bond’s yield to a benchmark yield curve. One widely used benchmark is an OIS (Overnight Index Swap) or RFR (Risk-Free Rate) curve, which is considered nearly risk-free. Unlike mortgage and corporate bond yield curves, OIS/RFR curves have minimal credit risk and liquidity premia. Such curves include the SOFR (Secured Overnight Financing Rate) curve in the U.S. market, the €STR (Euro Short-term Rate) in the Eurozone, and the STINA (SEK Overnight Index Swaps) in the Swedish market. To fulfill this criterion, we introduce the penalty term below: 
\begin{equation}
\label{eq:loss_trend}
\mathcal{L}_{\text{trend}} = \frac{1}{N} \textstyle\sum_{i=2}^N \left|\frac{\left(\hat{y}(t_i)-\hat{y}(t_{i-1})\right) - \left(y_{\text{\scriptsize OIS}}(t_i)-y_{\text{\scriptsize OIS}}(t_{i-1})\right)}{t_i - t_{i-1}}\right|.
\end{equation}
By compiling the above penalty terms as the total loss, we have:
\begin{equation}
\label{eq:loss_total}
    \mathcal{L} = \mathcal{L}_{\text{error}} + \gamma_1\mathcal{L}_{\text{smooth}} + \gamma_2\mathcal{L}_{\text{trend}},
\end{equation}
where $\gamma_1$ and $\gamma_2$ are hyperparameters indicating the weight of each penalty term in the overall loss. To train the network, we feed the cashflow dates of each bond to the network and first obtain the estimated spot yield rates. The corresponding discount factors are then computed using (\ref{eq:discount-yield}). Then, the loss $\mathcal{L}$ is calculated, and we update the network's parameters using backpropagation.
This process is iterated for all bonds in the training set, which is denoted as one epoch. We run this process for a certain number of epochs, which will be tuned as a hyperparameter.

\section{Experiments}
\label{sec:experiment}
\subsection{Data \& models}
In this study, we used mortgage bonds on the Swedish market. Each bond is represented by its market price, cashflow dates, cashflow amounts, and its maturity. Data are collected by the SEB Group's market risk team and consist of $\sim$60 bonds per day with a wide spread of maturities between a few weeks and more than $15$ years. In practice, it is expected that the estimated Swedish mortgage yield curve follows the trends of the SEKOIS curve, with extreme points occurring relatively close; therefore, we used the SEKOIS curve as the risk-free benchmark in (\ref{eq:loss_trend}). To argue the advantages of using our proposed NN-based model, we compare the estimations of the yield curve against the widely used parametric model NSS \cite{svensson1994estimating}, and the recent non-parametric KR model \cite{Filipovic2022stripping}.

\subsection{Hyperparameter selection}
\label{sec:hyperparam}
The neural network architecture used in our experiments consists of a single-layer network with three neurons and a \textit{tanh} activation function as per equation (\ref{eq:NN_model}), which we found to be sufficiently capable for the estimation task. To select the hyperparameters for our model, we examine the effect of varying the learning rate ($LR$), number of training epochs, and the parameters $\gamma_1$ and $\gamma_2$. 

We investigate the accuracy by computing the root-mean-square error (RMSE) between the bonds' YTM and the estimated yield at the corresponding maturities:
\begin{equation}
\label{eq:rmse_yield}
    RMSE_{\text{ytm}} = \sqrt{\frac{1}{M}\textstyle\sum_{j=1}^{M}{\left(y(t_j)-YTM_j\right)^2}},
\end{equation}
where $t_j$ is the time to maturity of the $j$-th bond.

\begin{table}[t]
\centering
\caption{RMSE$_{\text{ytm}}$ in a falling market, $\gamma_1=10^3$, $\gamma_2=10^4$}
\vspace{-10pt}
\begin{tabular}{@{}lccc@{}}
\toprule
 & $LR=10^{-7}$ & $LR=10^{-8}$ & $LR=10^{-9}$ \\
\midrule
epochs=$200$  & 0.2110 & 0.1972 & 4.2559   \\
epochs=$500$  & 0.2033 & \textbf{0.1790} &  3.6105  \\
epochs=$1000$ & 0.2929 & \textbf{0.1726} & 2.5843    \\
\bottomrule
\end{tabular}
\label{tab:hyper-8-2-2024}
\end{table}

\begin{table}[t]
\centering
\caption{RMSE$_{\text{ytm}}$ in a falling market, $LR=10^{-8}$, epochs$=10^3$}
\vspace{-10pt}
\begin{tabular}{@{}lcccc@{}}
\toprule
 & $\gamma_2=0$ & $\gamma_2=10^2$ & $\gamma_2=10^4$ & $\gamma_2=10^5$ \\
\midrule
$\gamma_1=0$                   & 1.9426      & 1.8472 & \textbf{0.1486} & 0.2982\\
$\gamma_1=10^3$        &  1.3825 & 1.2787 & \textbf{0.1726} & 0.2982\\
$\gamma_1=10^5$        & 0.2325 & 0.2319 & 0.2306 & 0.1942\\
\bottomrule
\end{tabular}
\label{tab:hyper-10-2024}
\end{table}

Figure~\ref{fig:hyperparam-merged} illustrates the results in a falling market scenario in which we estimate the yield curve using different choices of the hyperparameters. Note that the individual bonds are plotted in each figure in terms of their YTM, which differs from the spot yield shown on the Y-axis. Therefore, the best-fitting curve that passes through all individual bonds' YTMs does not necessarily lead to the best price accuracy. The associated RMSE$_{\text{ytm}}$ values are reported in Tables~\ref{tab:hyper-8-2-2024} and \ref{tab:hyper-10-2024}. Although increasing $\gamma_1$ leads to a smoother estimated curve (See Figure~\ref{fig:hyperparam-merged}-right), it does not necessarily reduce the RMSE, as evident from Table~\ref{tab:hyper-10-2024}. A larger $\gamma_2$ encourages the estimated curve to follow the SEKOIS benchmark more closely, often resulting in a more realistic shape. However, this may cause the model to deviate from market-observed prices, thereby increasing the RMSE$_{\text{ytm}}$.

\begin{figure*}[htbp]
    \centering
    \includegraphics[width=\linewidth]{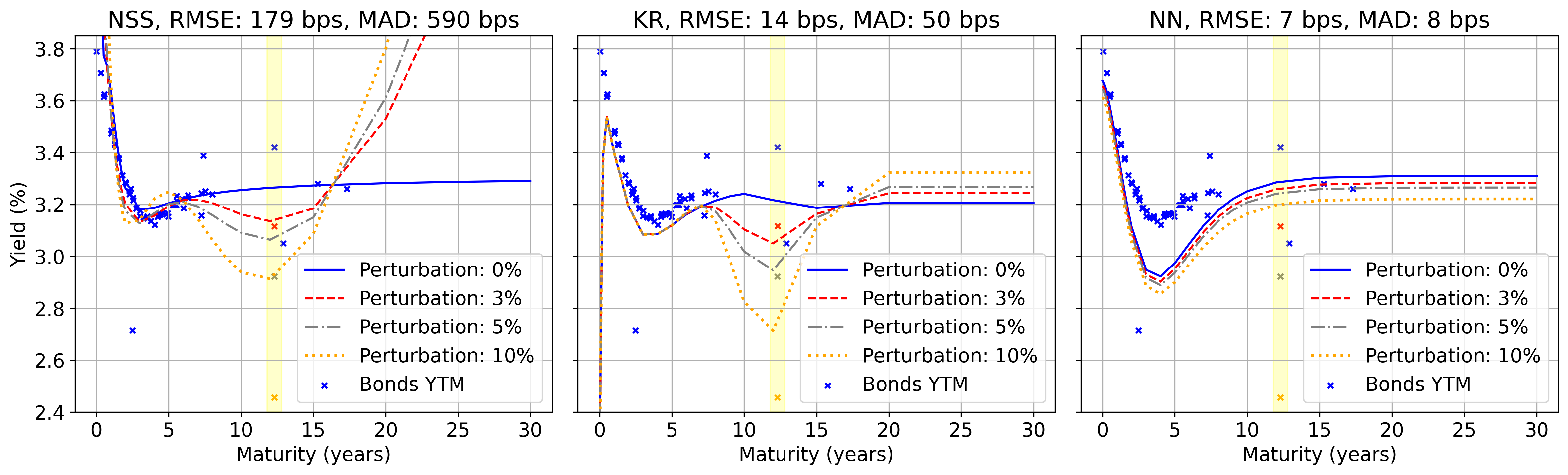}
    \caption{Robustness test for NSS, KR, and NN when perturbing the price of a bond with maturity 12.3Y by 3, 5, and 10\% increase.}
    \label{fig:robustness_single_merged}
\end{figure*}

The influence of $LR$ and the number of training epochs is summarized in Table~\ref{tab:hyper-8-2-2024}. The results suggest that at least 500 epochs are required for convergence, and a learning rate of $LR = 10^{-8}$ consistently yields the low RMSE$_{\text{ytm}}$ across scenarios. This is further supported by the behavior of the estimated curve with $LR = 10^{-8}$ and 1000 epochs in Figure~\ref{fig:hyperparam-merged}, where the curve remains above SEKOIS and exhibits stable behavior in both the short- and long-term segments. We ultimately select $LR = 10^{-8}$, 1000 epochs, $\gamma_1 = 10^3$, and $\gamma_2 = 10^4$ for the remainder of the experiment in this paper, unless otherwise specified. Although $\gamma_1=0$ results in a lower RMSE$_{\text{ytm}}$ as observed in Table~\ref{tab:hyper-10-2024}, we strike a balance between low RMSE$_{\text{ytm}}$ and desirable curve characteristics in our experiments.

\subsection{Robustness to outliers}
In this section, we compare different methods and evaluate the robustness of their estimated yield curves to the existence of outliers in the dataset. We first perturb the training data, either by changing the bond prices or removing bonds entirely, and then measure the sensitivity of each method using RMSE and maximum absolute difference (MAD) between the original unperturbed yield curve $y(t)$ and a reference yield curve $\tilde{y}(t)$ as follows:
\begin{equation}
\label{eq:rmse_robustness}
    RMSE_{\text{curve}} = \sqrt{\frac{1}{N}\textstyle\sum_{i=1}^{N}{(y(t_i)-\tilde{y}(t_i))^2}},
\end{equation}
\begin{equation}
    \label{eq:mad_robustness}
    MAD = \max_{t}|y(t)-\tilde{y}(t)|,
\end{equation}
where $\tilde{y}(t)$ is the perturbed yield curve. The maturity grid points $t_i$ that we use to compute RMSE$_{\text{curve}}$ are: 1D, 1W, 2W, 1M, 2M, 3M, 6M, 9M, 12M, 15M, 18M, 21M, 2Y, 3Y, 4Y, 5Y, 6Y, 7Y, 8Y, 9Y, 10Y, 12Y, 15Y, 20Y, 25Y, 30Y.

First, we visually compare the extent to which the yield curve is affected by different methods when the price of a single bond (with a maturity of approximately 12 years) increases by 3\%, 5\%, or 10\% on a given day. These scenarios test how well the models handle the presence of outliers, for example, when a callable bond is included in the dataset. For each case, we report perturbation RMSE$_{\text{curve}}$ and MAD in bps (basis points) for the case of 10\% perturbation and compare our NN with existing methods. The results are shown in Figure~\ref{fig:robustness_single_merged}. The NSS model performs poorly even under small perturbations, particularly at long maturities. Although it can be argued that there are fewer bonds at longer maturities, both the KR and our NN model manage to handle the long tail of the curve more effectively. In the comparison between the KR and NN models, it is evident that our model is superior in handling perturbations across different maturities, indicating that it is more robust when dealing with outliers and noise in the market.

\begin{figure*}[htbp]
    \centering
    \includegraphics[width=\linewidth]{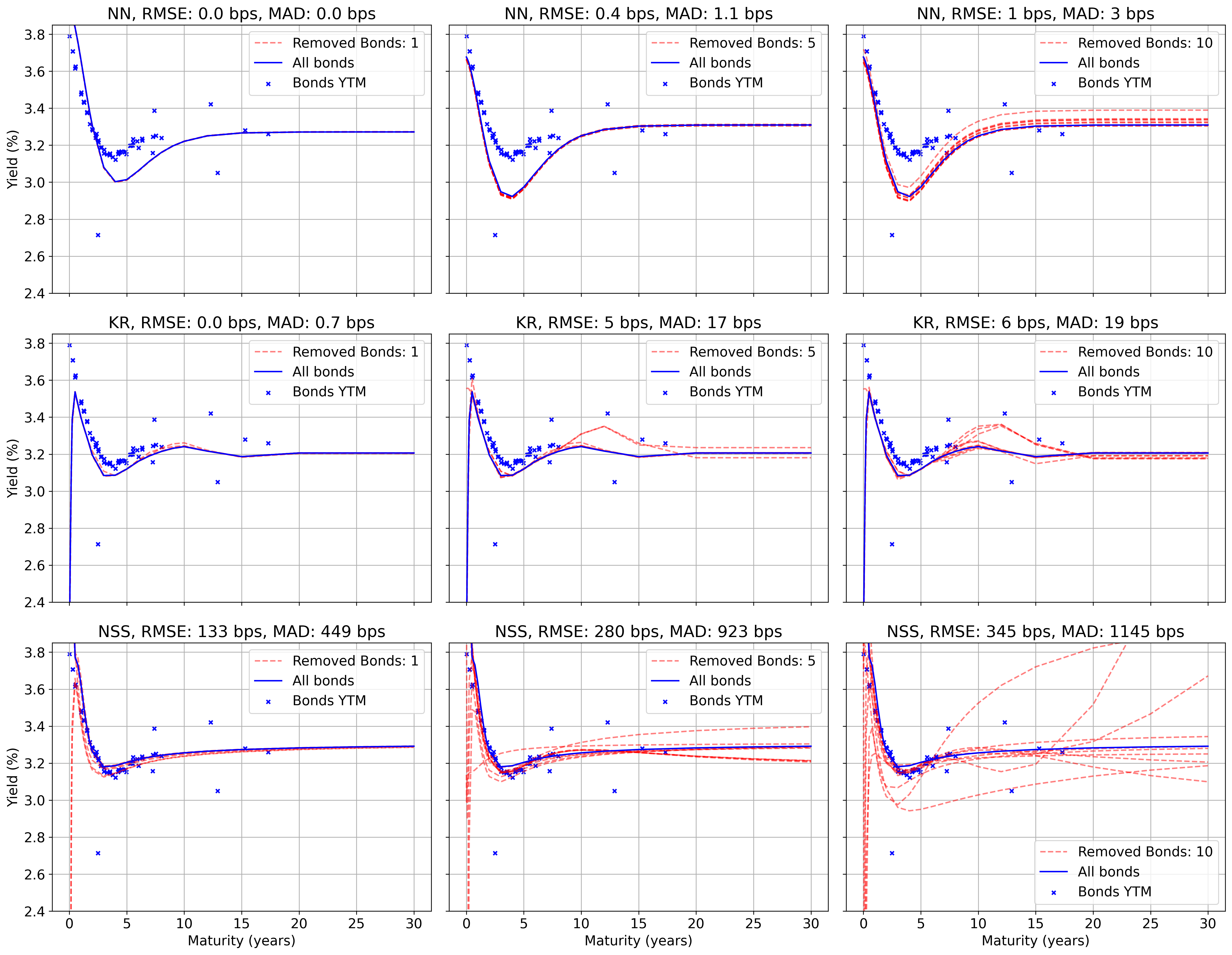}
    \caption{Robustness test for NSS, KR, and NN when randomly dropping 1, 5, and 10 bonds from the dataset for 10 MC simulations. The solid line is the yield curve estimated using all bonds. The dashed lines show the curves after randomly dropping bonds.}
    \label{fig:robustness_removal_merged}
\end{figure*}

Next, we compare the sensitivity of each method when removing one or more samples from the data. We randomly drop 1, 5, and 10 bonds from the data for a given day and compare RMSE$_{\text{curve}}$ and MAD for each case, averaged over 10 Monte Carlo (MC) simulations. The results are shown in Figure \ref{fig:robustness_removal_merged}, where we compare our NN model with KR and NSS (using the same random seed). The NSS model is highly sensitive to the removal of bonds from the estimation. Although the KR model is more robust than the NSS, perturbations in the curve at medium maturities affect the smoothness of the curve and are larger than those in our model, making the NN model more appealing in practice.

\subsection{Stability across days}
In this section, we demonstrate that NN estimations are less sensitive to changes in bond prices over time. Flexible models, such as high-degree splines, may chase idiosyncratic price movements rather than accurately reflecting accurate rate expectations. Bid-ask spreads in thin markets such as mortgage bonds introduce noise that standard models might misinterpret as rate changes. As a measure of the stability of the curve, over a span of 1 year in history, we calculate: 
\begin{enumerate}
    \item RMSE$_{\text{curve}}$ where $\tilde{y}(t)$ is the yield curve of previous day.
    \item \textit{Hit Rate} as the percentage of days where RMSE$_{\text{curve}}$$< 10$ bps. A hit rate of $>90\%$ is considered stable for liquid tenors.
    \item Daily yield rate estimation for maturities 6M, 2Y, and 10Y and comparing with SEKOIS rates.
\end{enumerate}

\begin{figure*}[htbp]
    \centering
    \includegraphics[width=\linewidth]{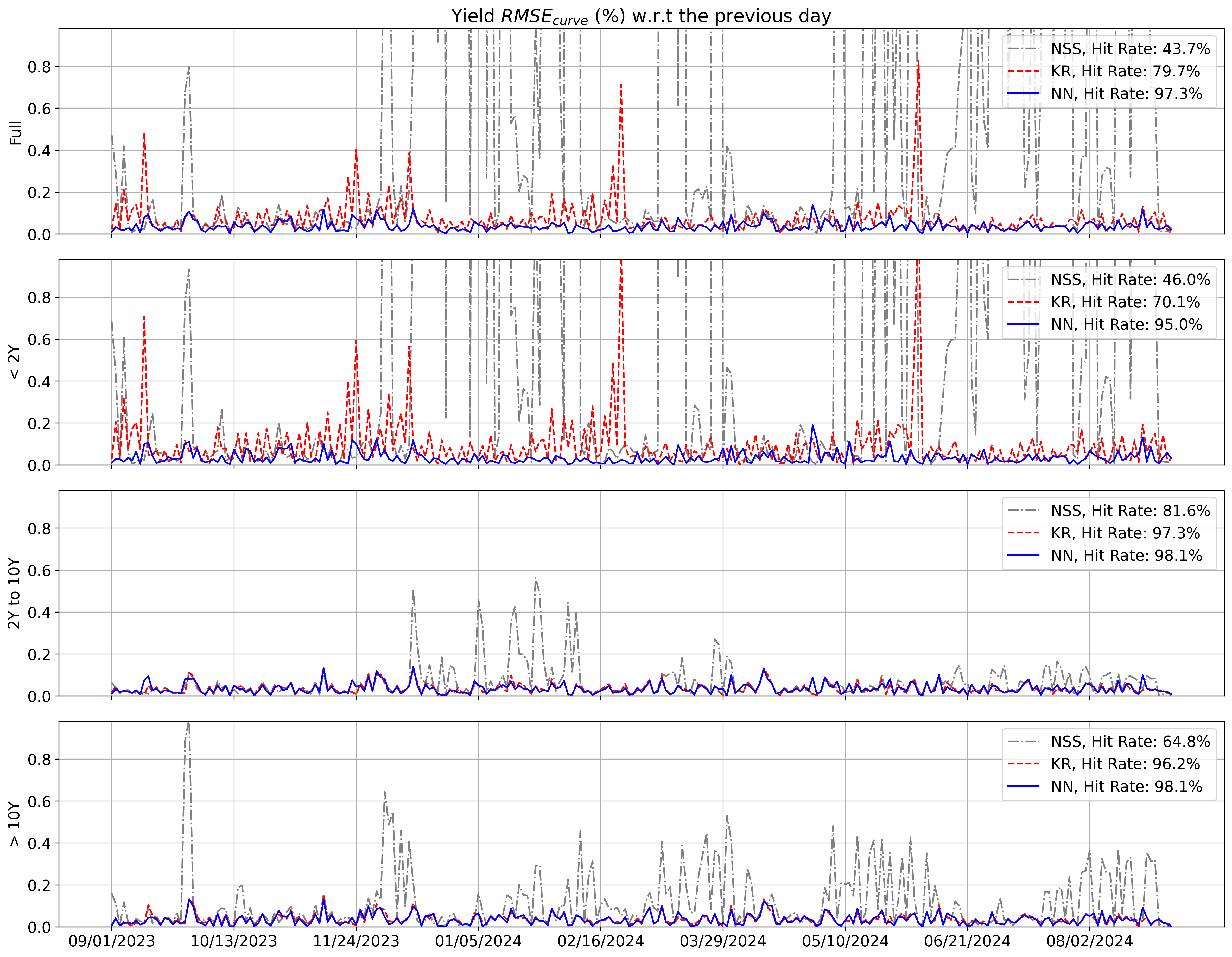}
    \caption{RMSE$_{\text{curve}}$ w.r.t to the previous day along with Hit Rate of RMSE < 10 bps over a period of 1 year.}
    \label{fig:stability_RMSE_ALL}
\end{figure*}

Figure~\ref{fig:stability_RMSE_ALL} shows the comparison between our NN model with KR and NSS in terms of the difference between today's curve and the previous day. We use different maturity buckets to illustrate the RMSE$_{\text{curve}}$ in different regimes. For a stable model, it is undesirable to observe large spikes in the calculated RMSE$_{\text{curve}}$. It is evident that the NN model exhibits smaller spikes and consistently higher hit rates compared to NSS across different maturity buckets, and it outperforms the KR model in hit rate in most experiments.

\begin{figure*}[htbp]
    \centering
    \includegraphics[width=\linewidth]{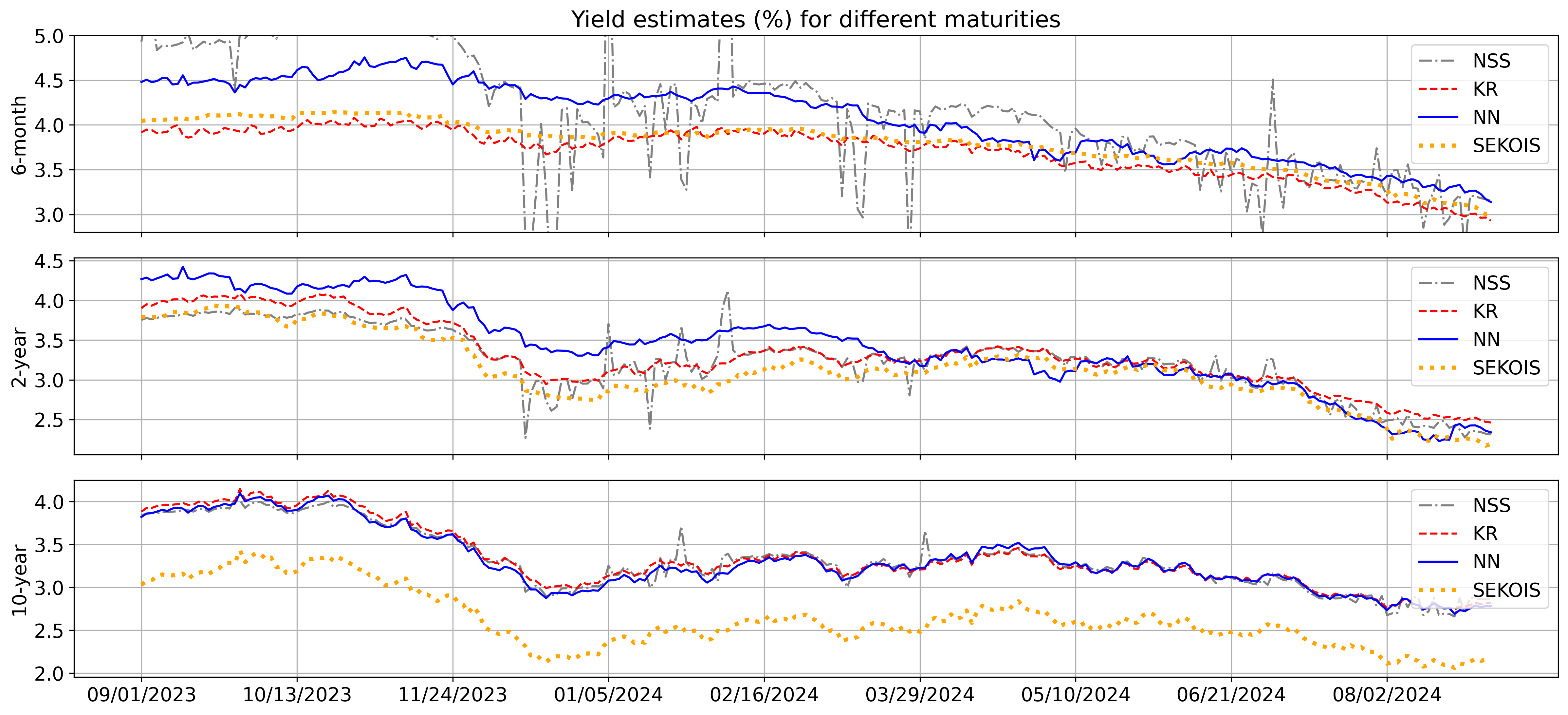}
    \caption{Stability test when estimating the yield of a specific maturity, namely, 6-month, 2-year, and 10-year, compare to the benchmark SEKOIS rate over a period of 1 year.}
    \label{fig:stability_fixed_day_estimate}
\end{figure*}

The comparisons in Figure~\ref{fig:stability_fixed_day_estimate} reveal that our model behaves more rationally compared to the risk-free rates for all three maturity examples. For shorter maturity (6M), the corresponding estimated rates using the KR and NSS models fall below the SEKOIS rate on many days, which is not justifiable. For 2Y and 10Y maturities, the models perform similarly, with the NSS model showing occasional spikes, which can be due to the high sensitivity of this model.

\begin{table*}[t]
\centering
\caption{RMSE$_{\text{ytm}}$ comparison across maturity buckets and market scenarios.}
\vspace{-10pt}
\begin{tabular}{@{}lcccccccccccc@{}}
\toprule
\multirow{2}{*}{\textbf{Model}} &
\multicolumn{4}{c}{\textbf{Flat (3/6/2020)}} &
\multicolumn{4}{c}{\textbf{Rising (1/6/2022)}} &
\multicolumn{4}{c}{\textbf{Falling (3/6/2024)}} \\
\cmidrule(lr){2-5} \cmidrule(lr){6-9} \cmidrule(lr){10-13}
& Full & $<$2Y & 2Y--10Y & $>$10Y
& Full & $<$2Y & 2Y--10Y & $>$10Y
& Full & $<$2Y & 2Y--10Y & $>$10Y \\
\midrule
NSS               & 0.2060 & 0.4163 & 0.0152 & 0.0525 & 0.1332 & 0.2254 & \textbf{0.0627} & 0.1585 & \textbf{0.1204} & 0.1488 & \textbf{0.0992} & 0.1846 \\
KR                & \textbf{0.0180} & \textbf{0.0151} & \textbf{0.0125} & \textbf{0.0426} & \textbf{0.0774} & \textbf{0.0572} & 0.0629 & 0.1542 & 0.1296 & 0.1427 & 0.1053 & 0.2390 \\
NN                & 0.1564 & 0.2494 & 0.1142 & 0.0871 & 0.2504 & 0.4519 & 0.1273 & 0.1451 & 0.1779 & \textbf{0.1058} & 0.1969 & 0.1589 \\
NN ($\gamma_1=0$) & 0.0882 & 0.1260 & 0.0692 & 0.0903 & 0.1431 & 0.2527 & 0.0646 & \textbf{0.1402} & 0.1512 & 0.1507 & 0.1519 & \textbf{0.1469} \\
\bottomrule
\end{tabular}
\label{tab:YTM-RMSE-LOO}
\end{table*}

\subsection{Smoothness vs flexibility trade-off}
\begin{figure*}[htbp]
    \centering
    \includegraphics[width=\linewidth]{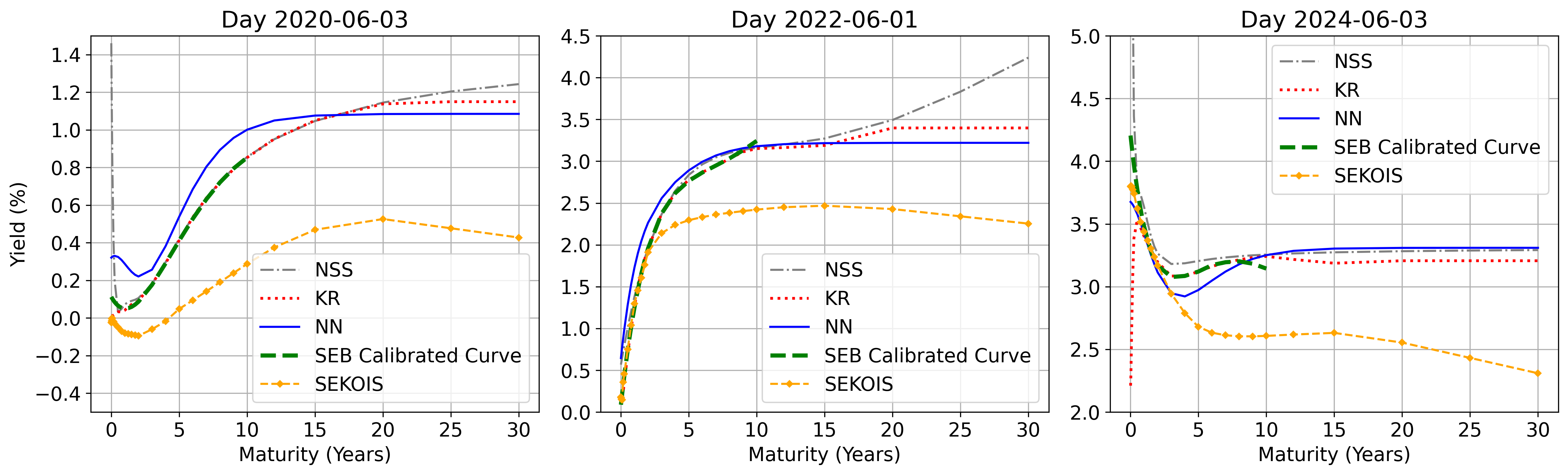}
    \caption{Comparing an example of LOO estimated yield curve for three different days representing a flat, rising, and falling market condition. Dropped bond is chosen from the bucket of <2Y maturities.}
    \label{fig:smothness_comparison}
\end{figure*}

In this section, the two objectives of "accuracy" and "economical reasonableness" in Section~\ref{sec:neural_net} are revisited in more extensive scenarios.
We consider three different days in history with noticeably different shapes of the SEKOIS curve (market scenarios). To experiment with the out-of-sample performance, we exclude one bond in the training and compute the yield error and price error using the excluded sample. Then leave-one-out (LOO) yield RMSE$_{\text{ytm}}$ is calculated on average over 10 Monte Carlo simulations as a measure of pricing accuracy, while the behavior of our estimated curve is compared against the SEKOIS curve and an in-house calibrated curve (at SEB Group) for three different days and compared against existing methods such as KR and NSS (see Figure~\ref{fig:smothness_comparison}). 
In the example of the rising market (2020-06-03), the KR and NN models have a justified spread relative to the SEKOIS curve and SEB-calibrated curves, while the NSS model performs poorly at short maturities. In the flat market example (2022-06-01), the NN model has the advantage of estimating a smoother curve than the KR model, and more justified rates at the long tail. Finally, in the falling market example (2024-06-03), the NN model shows better estimations compared to KR and NSS, as the KR model falls below the SEKOIS curve at short maturities and the NSS model exhibits unjustifiably increasing rates at the long end. Having a smooth behavior and justified with respect to SEKOIS comes at the price of losing accuracy as indicated in Table~\ref{tab:YTM-RMSE-LOO}. When we regularize the loss function less in the NN model ($\gamma_1=0$), the RMSE$_{\text{ytm}}$ drops and falls in the same range as other estimators.

\section{Conclusion}
\label{sec:conclusion}
We demonstrated that utilizing neural networks for yield curve estimation can provide a more robust and stable estimate, particularly in smaller and relatively less liquid markets, such as the Swedish mortgage bond market. We compared our results against NSS and KR in various market conditions and achieved a smoother curve in all scenarios. This, however, is achieved at the cost of sacrificing the accuracy of the curve in terms of yield RMSE on LOO samples. In this way, NNs provide a framework that allows analysts to tune the model to their specific needs and balance the trade-off between accuracy and economic reasonableness as they see fit. Optimization of the NN architecture is a potential future direction for improving the RMSE of our model. Incorporating temporal data to enable yield curve forecasting using neural networks is another promising area for future research. 

\bibliographystyle{ACM-Reference-Format}
\bibliography{references}

\end{document}